\pdfoutput=1

\documentclass[11pt]{article}

\usepackage{acl}

\usepackage{multirow}
\usepackage{graphicx}
\usepackage{pifont}
\usepackage{amssymb}
\usepackage{caption, subcaption}
\usepackage{array}
\usepackage{booktabs}
\usepackage{times}
\usepackage{latexsym}
\usepackage{comment}
\usepackage[T1]{fontenc}

\usepackage[utf8]{inputenc}

\usepackage{microtype}

\usepackage{balance}
\usepackage{comment}
\usepackage{url}
\usepackage{graphicx}
\usepackage{amsmath}
\usepackage{enumitem}
\usepackage{booktabs}
\usepackage{array}
\usepackage{caption, subcaption}
\usepackage{balance}

\usepackage{color, colortbl}
\usepackage{multirow}
\newcolumntype{P}[1]{>{\centering\arraybackslash}p{#1}}
\definecolor{Gray}{gray}{0.95}
\definecolor{Gray1}{gray}{0.9}
\definecolor{Gray2}{gray}{0.85}
\definecolor{Gray3}{gray}{0.8}
\definecolor{Red1}{rgb}{1,0.78,0.65}
\definecolor{Red2}{rgb}{1,0.88,0.78}
\definecolor{Red3}{rgb}{1,0.7,0.6}
\definecolor{Red4}{rgb}{1,0.6,0.5}
\definecolor{Red5}{rgb}{1,0.93,0.83}
\definecolor{Green1}{rgb}{0.95,0.97,0.75}
\definecolor{Green2}{rgb}{0.9,0.95,0.7}
\definecolor{Green3}{rgb}{0.8,0.85,0.6}
\definecolor{Yellow}{rgb}{1,1,0.9}
\newcolumntype{r}{>{\columncolor{Gray}}c}
\newcolumntype{s}{>{\columncolor{Gray1}}c}
\newcolumntype{t}{>{\columncolor{Gray2}}c}
\newcolumntype{u}{>{\columncolor{Gray3}}c}
\newcolumntype{e}{>{\columncolor{Red}}c}
\newcolumntype{f}{>{\columncolor{Blue}}c}
\newcolumntype{y}{>{\columncolor{Yellow}}c}

\title{Predicting Evoked Emotions in Conversations}

\author{Enas Altarawneh \\   York University \\
\texttt{enas@eecs.yorku.ca}  \\ \And 
  Ameeta Agrawal \\
  Portland State University\\
\texttt{ameeta@pdx.edu} \\ \vspace{-15pt}
  \\\AND
  Michael Jenkin \\
  York University\\
\texttt{jenkin@eecs.yorku.ca} \\
  \\\And
  Manos Papagelis \\ 
  York University\\ \texttt{papaggel@eecs.yorku.ca} \\
  }

\begin{document}

\maketitle
\begin{abstract}
Understanding and predicting the {\em emotional trajectory} in {\em multi-party multi-turn conversations} is of great significance. Such information can be used, for example, to generate empathetic response in human-machine interaction or to inform models of pre-emptive toxicity detection. In this work, we introduce the novel problem of {\em Predicting Emotions in Conversations} (PEC) for the next turn ($n+1$), given combinations of textual and/or emotion input up to turn $n$. We systematically approach the problem by modeling three dimensions inherently connected to evoked emotions in dialogues, including (\textbf{i}) {\em sequence modeling}, (\textbf{ii}) {\em self-dependency modeling}, and (\textbf{iii}) {\em recency modeling}. These modeling dimensions are then incorporated into two deep neural network architectures, a {\em sequence model} and a {\em graph convolutional network model}. The former is designed to capture the sequence of utterances in a dialogue, while the latter captures the sequence of utterances and the network formation of multi-party dialogues. We perform a comprehensive empirical evaluation of the various proposed models for addressing the PEC problem. The results indicate (\textbf{i}) the importance of the self-dependency and recency model dimensions for the prediction task, (\textbf{ii})  the quality of simpler sequence models in short dialogues, (\textbf{iii})  the importance of the graph neural models in improving the predictions in long dialogues.
\end{abstract}

\section{Introduction}


Automatic emotion recognition in conversations has numerous applications and has been extensively studied, typically as the process of estimating emotions of a specific utterance. But utterances are rarely given in isolation and they are rather part of a conversation. A more challenging, but desirable in many applications, task is the ability to predict the emotion trajectory of a conversation before the actual (future) utterances become available to the model. Towards this end, we introduce the novel problem of {\em Predicting Evoked Emotions in Conversations (\textsc{PEC})\/}.
Given a sequence of utterances in a turn-taking conversation, we want to predict the likelihood of certain emotion(s) being expressed by a speaker over the next turn. An example conversation coming from a two-speaker conversation dataset, \textsc{DailyDialog}, is presented in Table~\ref{tab:example}.



\begin{table}[t!]
\centering
\resizebox{\columnwidth}{!}{

\begin{tabular}
{p{0.8cm}p{0.3cm}p{5cm}c}
\toprule
turn & user & text & emotion\\  
\midrule
$n-2$  & $\mathcal{A}$ & {\em ``Is everything alright?"} &  neutral \\
$n-1$ & $\mathcal{B}$ & {\em  ``No, the steak is not very fresh." } &  anger\\
$n$ & $\mathcal{A}$ & {\em  ``Oh! Sorry to hear that."} & sadness\\
$n+1$ & $\mathcal{B}$ & \hrule & \textbf{\textcolor{red}{???}}\\
\bottomrule
\end{tabular}
}
\caption{A sample conversation from \textsc{DailyDialog} showing text utterances and their emotion labels. Given $n$ turns, the task is to predict the emotion at turn $n+1$ (\textcolor{red}{anger}, in this case).}
\label{tab:example}
 \vspace{-4pt}
\end{table}

\begin{figure}[t!]
     \centering
     \begin{subfigure}{0.235\textwidth}
         \centering
         \vspace{-5pt}
         \includegraphics[width=\textwidth]{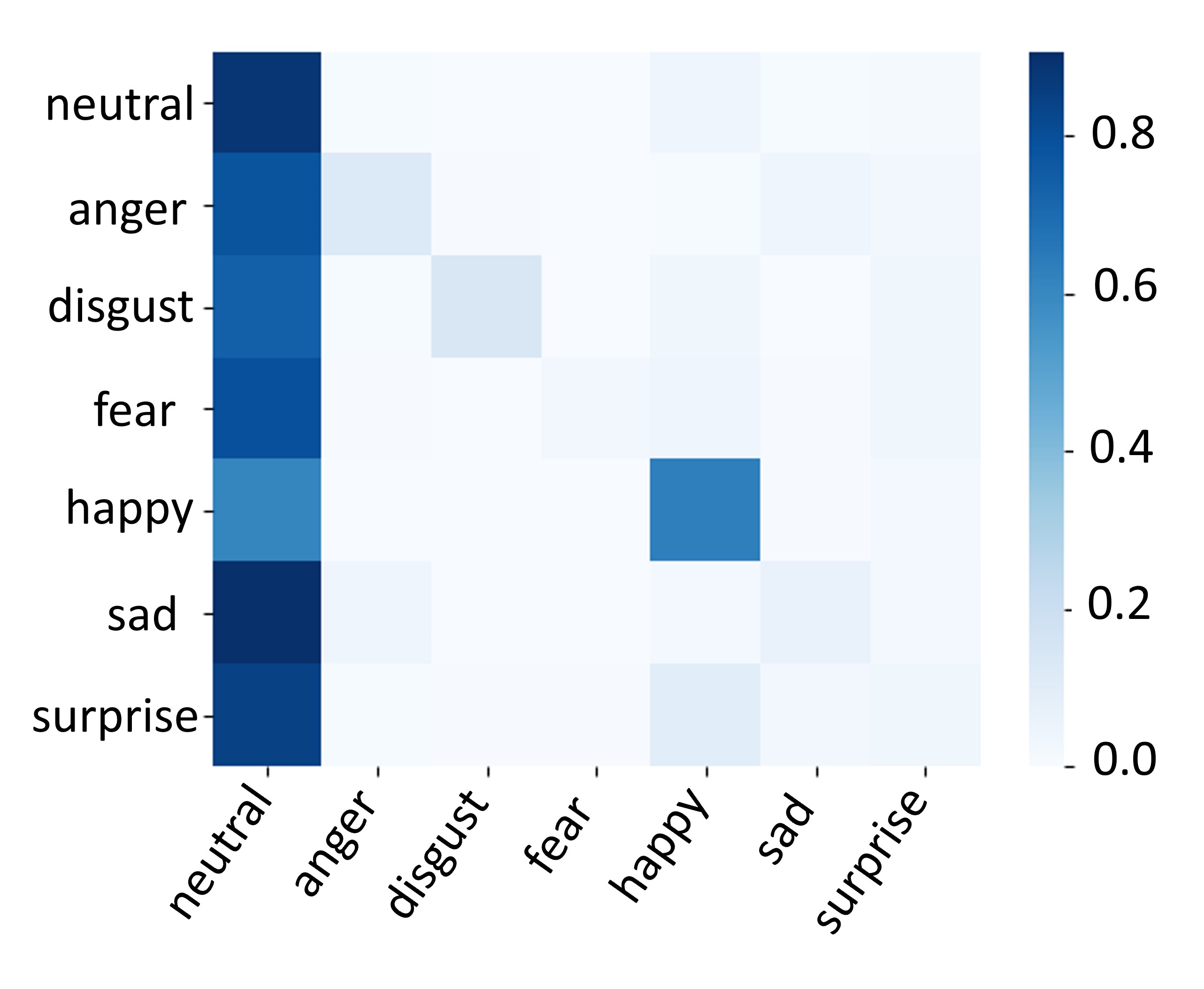} 
         \caption{$P(e_{n+1}|e_n)$}
         \label{fig:cm1}
     \end{subfigure}
     \begin{subfigure}{0.235\textwidth}
         \centering
         \vspace{-5pt}
         \includegraphics[width=\textwidth]{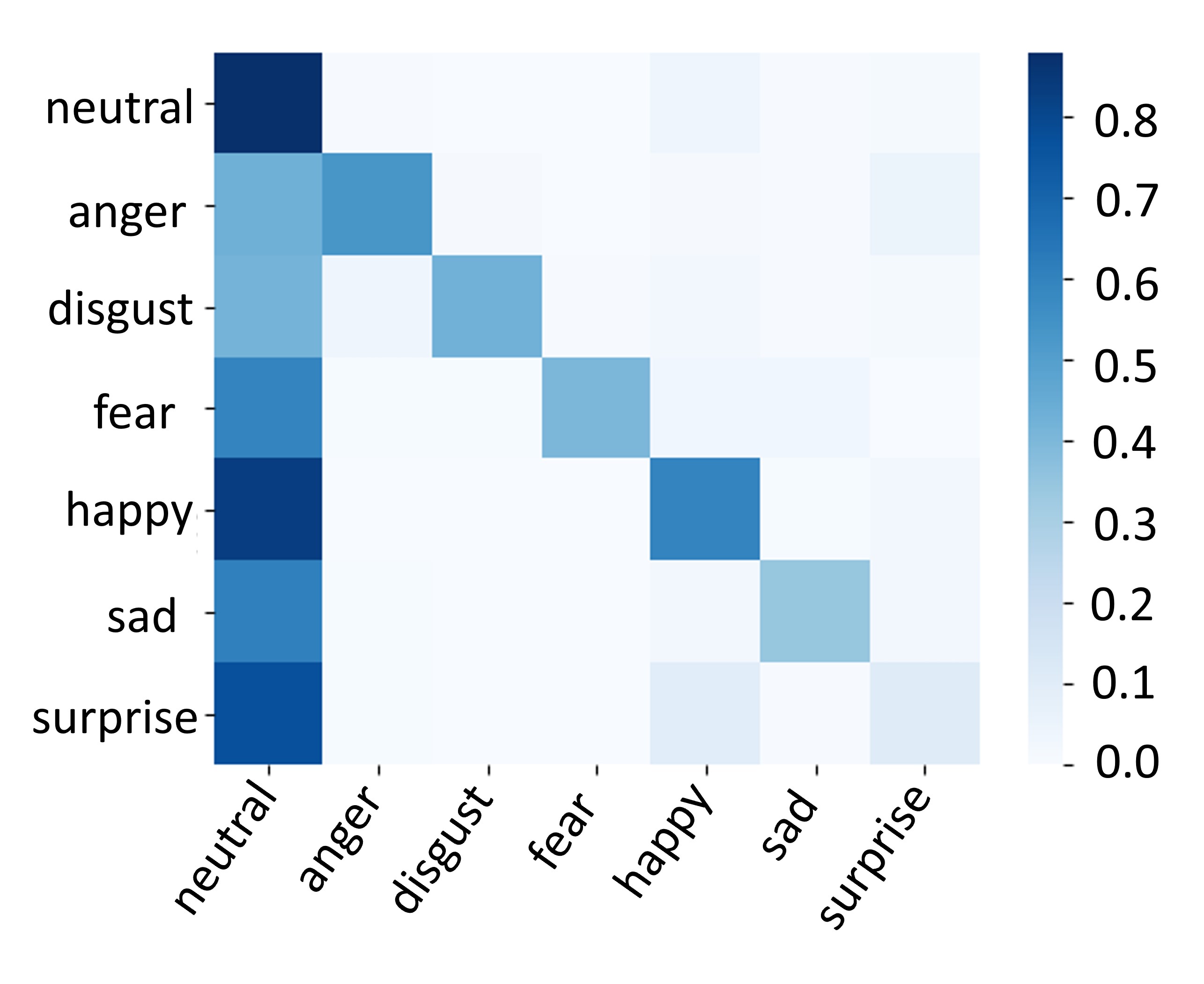} 
         \vspace{-10pt}
         \caption{$P(e_{n+1}|e_{n-1})$}
         \label{fig:cm2}
     \end{subfigure}
\caption{Transition matrix showing the transition probabilities of one emotion to another. $e_{n+1}$ and $e_{n-1}$ come from the same user; $e_n$ pertains to another user. }
\label{fig:cm}
\vspace{-0.5cm}
\end{figure}



There is a wide range of applications for which such a forecasting model can be useful, including forecasting the emotional trajectory of a conversation between a machine and human agent or pre-emptively detecting hate speech in social forums \cite{horne2017identifying, davidson2017automated, martins2018hate,ren2020review,poletto2020resources}. 
To further motivate the problem and challenge, Figure~\ref{fig:cm} shows the transition matrix of emotions in  \textsc{DailyDialog} between any two turns. In Figure~\ref{fig:cm1}, we observe that for the majority of the cases, there is a high probability of transitioning to a {\em neutral} state when the two turns pertain to {\em two} different users, except in the case of {\em happy} which is more likely to be mimicked by the other party. Figure~\ref{fig:cm2}, on the other hand, suggests that emotion consistency is typically maintained when the two turns pertain to the {\em same} user (i.e., self-dependency). Interestingly, we notice that {\em surprise} is likely to transition to {\em happy} over the next turn. These inconspicuous insights of our preliminary analysis motivated us to further explore this line of research, and make the following contributions:
%
%
\begin{itemize}
    \item we introduce the novel problem of {\em Predicting Evoked Emotions in Conversations (\textsc{PEC})\/}.
    \item we systematically study the {\em modeling dimensions} of the problem, including aspects of (\textbf{i}) sequence modeling, (\textbf{ii}) self-dependency modeling, and (\textbf{iii}) recency modeling.
    \item we propose sensible deep neural network architectures, including {\em a sequence model} and a {\em graph convolutional network model} that incorporate the three modeling dimensions.
    \item we perform an extensive empirical evaluation of the proposed models across four datasets and provide a thorough report of the analysis that can inform adoption of the model in real scenarios and diverse applications.
    \item we (aim to) make source code publicly available to encourage reproducibility.
\end{itemize}

The remainder of the paper is organized as follows. Section \ref{sec:related} reviews the related work. The technical problem of interest in this paper is presented in Section \ref{sec:problem}. The modeling dimensions are discussed in Section \ref{sec:modeldimensions} and our proposed models are introduced in Section \ref{sec:models}. Section~\ref{sec:experiments} presents an experimental evaluation of the different models, and the conclusions are presented in Section~\ref{sec:conclusions}.
\section{Related Work}
\label{sec:related}


The task of emotion recognition in conversation (ERC) which detects the emotion at turn $n$ in a conversation given a conversation history from turn $1$ to turn $n$ has received significant attention \cite{ghosal2019dialoguegcn,poria2019emotion}. Here, we shift our attention towards a novel task, i.e., \textsc{PEC}, by developing models for \textbf{predicting emotions at turn $n+1$ given data up to only turn $n$}. Our work shows that speaker information is of significance. One ERC that addresses the need for speaker related information is DialogueGCN \cite{ghosal2019dialoguegcn}. In this work, we create an extended version of DialogueGCN to address our problem. 

Psychological studies show that humans create mental models of emotion transitions and can predict others' emotions up to two transitions into the future with an above-chance accuracy \cite{thornton2017mental}. \citet{zhou2017emotional} propose an emotional chatting machine which can react to the post with a required emotion using a seq2seq-based affective conversational model that takes as an input a prompt and the desired emotion category of the response, and produces a response, while \citet{huang2018automatic} implement several strategies to embed emotion into seq-to-seq models. \citet{zhou2017mojitalk} incorporate reinforcement learning into emotional response generation based on a large dataset labeled with emojis. \citet{colombo2019affect} design an affect sampling method to force the neural network to generate emotionally relevant words. \citet{kong2019adversarial} propose a method for neural dialogue response generation that allows not only generating semantically reasonable responses according to the dialogue history, but also explicitly controlling the sentiment of the response via sentiment labels. \citet{asghar2020generating} develop affect-aware neural conversational agents, which produce emotionally aligned responses to prompts. Although these studies show the possibility of generating a response capable of conveying an emotion, the approach is limited in that the emotion of the response should be determined manually by the user.

One natural application of such a predictive model is in the area of empathetic response generation where existing strategies either mimic previous emotion, require pre-determined emotion signals \cite{zhou2017emotional,huang2018automatic,zhou2017mojitalk,colombo2019affect} or jointly model emotion prediction and response generation via Conditional Variational Auto-Encoders (CVAEs) \cite{lubis2018eliciting,asghar2018affective,asghar2020generating,chen2019neural,gu2019towards}. However, CVAEs do not provide an interpretable model of emotions which could be used to derive insights about emotions in conversations as well as inform future models. In addition, such prediction can also be useful in the task of pre-emptive toxicity detection or the problem of detecting early indicators of antisocial discourse, which has become a pertinent research topic. \citet{zhang-etal-2018-conversations} studied linguistic markers for politeness strategies while \citet{brassardgourdeau2020using} extended that line of work by including sentiment information.


As human emotions are inherently ambiguous, a probabilistic distribution over the emotion categories may seem like a more reasonable representation. Emotional profiles (EPs) provide a time series of segment-level  labels to capture the subtle blends of emotional cues present across a specific speech utterance \cite{Mao2020}. Such profiles can be used for affect-sensitive human-machine interaction systems. Well-designed emotion recognition systems have the potential to augment such systems \cite{5585726}. Our work stresses the need for a more comprehensive understanding of emotional profiles that go beyond the utterance-level sequences and incorporate user specific signals. 

\section{Problem Statement}
\label{sec:problem}


In this section, we formally define the problem of {\em Predicting Emotion in Conversation (PEC)}. Let $\mathcal{C}=\{ \langle t_1,e_1\rangle, ..., \langle t_n,e_n\rangle\}$ denote a conversation of $n$ turns, where $t_i=\{w_1,...,w_m \}$ represents the sequence of words uttered at turn $i$ and $e_i \in \mathcal{E}$ where $\mathcal{E}$ is a finite set of emotion categories. We consider a {turn} to be a continuous and uninterrupted utterance/portion of a conversation by one user, and assume pre-existing databases of such conversations.  Given some  conversation $\mathcal{C}$ consisting of a specific sequence of labeled utterances up to and including time $n, n\geq1$, the task is to predict the emotion at the {\em next\/} turn, $e_{n+1}$.
\section{Modeling Dimensions}
\label{sec:modeldimensions}




We systematically approach the \textsc{PEC} problem by modeling three dimensions inherently connected to evoked emotions in dialogues, including (\textbf{i}) {\em sequence modeling}, (\textbf{ii}) {\em self-dependency modeling}, and (\textbf{iii}) {\em recency modeling}. We further elaborate on each of them in the following subsections.

\subsection{Sequence Modeling}
Given a conversation $\mathcal{C}$ of $n$ turns, our goal is to predict the emotion of the next turn $e_{n+1}$. We consider three cases that treat the sequence of conversation turns as a {\em sequence of emotions}, {\em sequence of texts} or {\em sequence of (emotion, text) pairs} as follows.

\smallskip\noindent\textbf{Sequence of emotions}: We use the sequence of emotion labels $\mathcal{C}^{\langle e \rangle}=\{ e_1,e_2,...,e_n\}$, $e \in \mathcal{E}$  to predict $e_{n+1}$. $e_i$ is a one-hot encoded vector of size $1 \times |\mathcal{E}|$, with each dimension representing one of the emotion classes $\mathcal{E}$.

\smallskip\noindent\textbf{Sequence of texts}: We utilize the sequence of text utterances $\mathcal{C}^{\langle t \rangle}=\{ t_1,t_2,...,t_n\}$ to predict $e_{n+1}$.

\begin{figure*}[h]
   \includegraphics[width= 16cm]{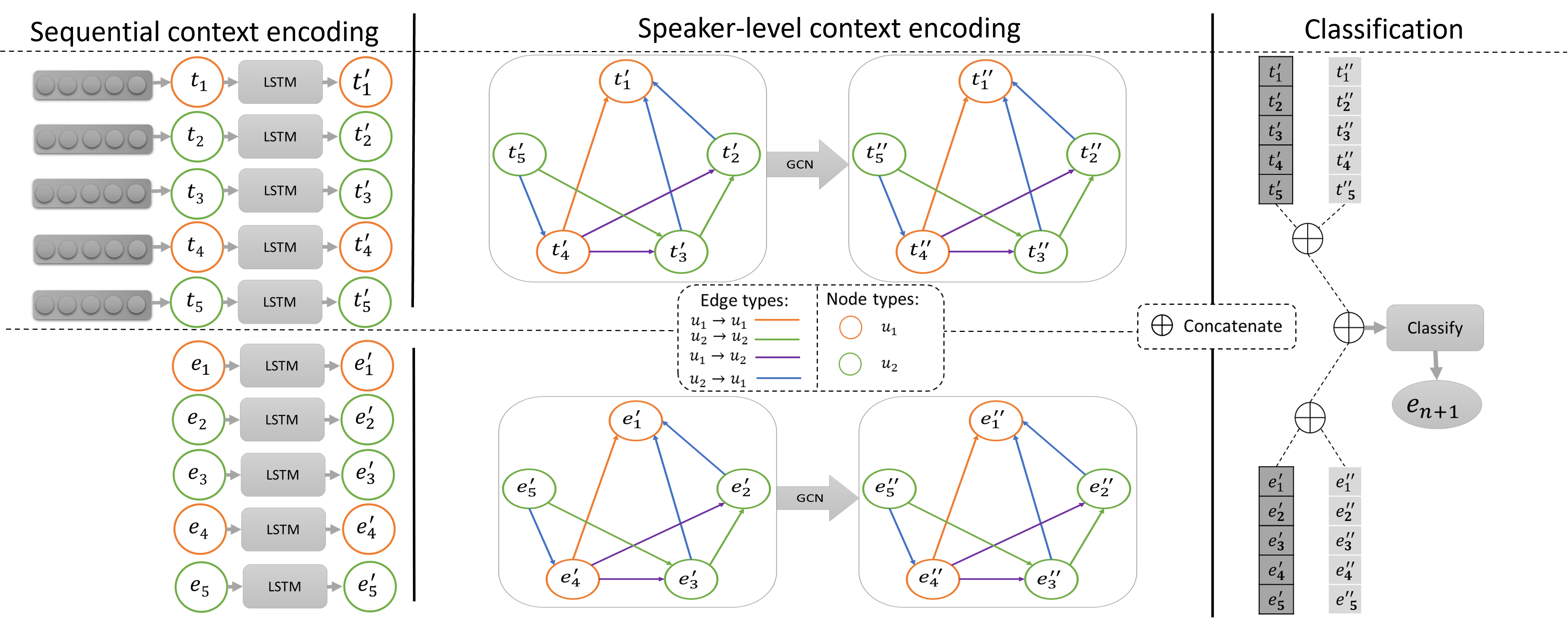}
   \caption{Overview of the proposed DGCN-PEC model architecture.}
   \label{GCN}
    \vspace{-0.5cm}
\end{figure*}

\smallskip\noindent\textbf{Sequence of (emotion, text) pairs}: 
We utilize the sequence of (text, emotion) pairs of utterances $\mathcal{C}=\{ \langle t_1, e_1 \rangle, \langle t_2, e_2 \rangle,..., \langle t_n, e_n \rangle\}$ to predict $e_{n+1}$. We construct the sequence of emotion labels $\mathcal{C}^{\langle e \rangle}=\{ e_1,e_2,...,e_n\}$ and the sequence of text utterances $\mathcal{C}^{\langle t \rangle}=\{ t_1,t_2,...,t_n\}$ and feed them separately into the next layer.

\subsection{Self-dependency Modeling}
The sequence models presented so far are agnostic to the the identity of the speaker. However, the nature of an evoked emotion might be dependent to the way a specific speaker converses. Does the model rely mostly on the utterances of the speaker being modeled ({\em self-dependency}), or on other participants in the conversation ({\em other-dependency})?
To address this question, we design and train variants of the sequence models that explore the nature of self-dependency and other-dependency to evoked emotions. Let a group of $m$ people $\mathcal{U}=\{u_1, u_2,\ldots,u_m\}$ that participate in a group conversation. Now, given a conversation $\mathcal{C}$ and a specific speaker $u \in \mathcal{U}$ (representing {\em self}), we can define two sequences of utterances $\mathcal{C}_{u}$ and $\mathcal{C}_{\mathcal{O}}$, such that $\mathcal{C}_{u}$ represents all utterances of $u$ and $\mathcal{C}_{\mathcal{O}}$ represents all utterances coming from any of the other speakers $\mathcal{O} = \mathcal{U} \setminus u$. Note that $\mathcal{C} = \mathcal{C}_{u} \cup \mathcal{C}_{\mathcal{O}}$. Similarly, after running our prediction models (see Section \ref{sec:models}) on the input conversation $\mathcal{C}$ and prior to final classification, we obtain a representation of the conversation $\mathcal{C'} = \mathcal{C'}_{u} \cup \mathcal{C'}_{\mathcal{O}}$.

\subsection{Recency Modeling}
The sequence models presented so far assume that all the $n$ turns of a conversation $\mathcal{C}$ inform the sequence model. However, the nature of an evoked emotion might be triggered by recent turns of the conversation \cite{fridhandler1982temporal}. Does the model rely on all utterances of the conversation or focus on the more recent ones? To address this question, we design and train variants of the sequence models that explore the {\em temporal dimension} of evoked emotions. Formally, we define the length $w$ of a {\em temporal look back window} that controls how far the sequence extends into the past upon which estimation relies explicitly.

\section{Models}
\label{sec:models}

In this section, we design and develop sensible deep neural network architectures that incorporate the three modeling dimensions, including a {\em sequence model} and a {\em graph convolutional network model}. 
We further elaborate on each of them next.

\subsection{BiLSTM-PEC}
To capture the sequence of utterances in a dialogue we introduce \textbf{BiLSTM-PEC}, a BiLSTM-based model. LSTM-based models are well-suited to classifying and making predictions based on sequence data and are known to outperform traditional recurrent neural networks (RNNs). In addition, Bidirectional LSTMs (BiLSTMs) enable additional training by traversing the input data twice (i.e., they exploit future and history context together at once). BiLSTM-based modeling offers better predictions than regular LSTM-based models, making it a sensible choice for our PEC problem.

Back to our problem's semantics, {\em text sequences} of each utterance are pre-processed (removal of punctuation and stopwords, lower-casing, and lemmatization) and converted into a vector representation using GloVe embeddings \cite{pennington2014glove}. {\em Emotion sequences} are converted into a vector representation before provided to the neural network. Finally, regarding {\em (emotion, text) pair sequences},  emotion and text sequences are treated separately (as if in isolation) and they are concatenated at a later stage, just before the final classification layer.
To exploit $u$'s self-dependency in predicting the emotion of its next utterance we train a model on the sequence of utterances $\mathcal{C}_{u}$ (self-dependency) and classify using $\mathcal{C'}_{u}$ . To explore the influence of other speakers' information in predicting the emotion of $u$'s next utterance, we train a model on the sequence of utterances $\mathcal{C}_{\mathcal{O}}$ (other-dependency) and classify using $\mathcal{C'}_{\mathcal{O}}$.  

In terms of implementation, vectors (word embeddings) are provided into a Time Distributed layer followed by a Flatten layer before passing them to a BiLSTM with attention layer. \verb|ReLu| is used as the dense layer activation function and \verb|softmax| as the output layer's activation function.

\subsection{DGCN-PEC}
To capture the sequence of utterances and the network formation of multi-party dialogues we introduce \textbf{DGCN-PEC}, an extension of the DialogueGCN model \cite{ghosal2019dialoguegcn}, designed to address the PEC problem. DialogueGCN is an ERC classifier that given conversation text utterances as input (similar to our problem), classifies each utterance to one of a given set of emotion classes. DialogueGCN incorporates speaker information by modeling multi-party conversations as a graph, where nodes represent utterances and edges (connecting two utterances) represent the speaker type relationship. Further details on the specifics of DialogueGCN can be found in \citet{ghosal2019dialoguegcn}.

In contrast to the base DialogueGCN model, our DGCN-PEC model gets as input a combination of text and/or emotion signals of conversation utterances, and predicts the emotion class of the next turn, $e_{n+1}$, in a conversation of size $n$, which is a different task than the one DialogueGCN is designed for. A high level architecture of our DGCN-PEC model is provided in Figure~\ref{GCN}. The input of the model is a one-hot emotion vector and text vectors (GloVe embeddings) of the conversation. Specifically, we use BiLSTM as the base model in the sequential context encoding and use the same speaker-level context encoding as DialogueGCN does. The speaker-level context encoding creates a graph of utterances. The nodes are instantiated with features extracted using the sequential context encoding. The utterances are connected through edges that reflect speaker relationships. The utterance features are transformed using a graph convolutional network. The sequential and speaker level context encoding output features are concatenated prior to classification. Similarly to BiLSTM-PEC, the {\em (emotion, text) pair sequences} are processed separately (as a sequence of emotions and  a sequence of texts), and their output is concatenated before the final classification layer.
To exploit $u$'s self-dependency in predicting the emotion of its next utterance we train a model on the sequence of utterances $\mathcal{C}$ (the entire conversation, so as to create a graph of the utterances), extract the $\mathcal{C'}_{u}$ part of the conversation from $\mathcal{C'}$, and classify on $\mathcal{C'}_{u}$ (self-dependency). Similarly, to explore the influence of other speakers' information in predicting the emotion of $u$'s next utterance, we train a model on the sequence of utterances $\mathcal{C}$, extract the $\mathcal{C'}_{\mathcal{O}}$ part of the conversation from  $\mathcal{C'}$, and classify on $\mathcal{C'}_{\mathcal{O}}$ (other-dependency). 

In terms of implementation, when the DGCN-PEC constructs the conversation graph in the speaker-level context encoding, there are two parameters, $pw$  (past window) and $fw$ (future window), which control how far in the past or future in the conversation to look at when creating the edges between the utterance nodes. Our empirical analysis on varying values of these parameters showed that $pw=3$ and $fw=0$ provide better results. We therefore employ these values in the experiments.

\subsection{Final Models}
Based on the aforementioned modeling dimensions and model architectures we define the following look back (LB) models.

\smallskip\noindent\textbf{\textit{w}LB}: A sequence model that is agnostic to the identity of the speaker and considers a temporal window of length $w$ in $\mathcal{C}$.

\smallskip\noindent\textbf{\textit{w}SLB}: A sequence model that considers self-dependency on speaker $u \in \mathcal{U}$ (i.e., only utterances of speaker $u$ are used) and a temporal window of length $w$ in $\mathcal{C}_{u}$ for \textbf{BiLSTM-PEC} or $\mathcal{C'}_{u}$ for \textbf{DGCN-PEC}.

\smallskip\noindent\textbf{\textit{w}OLB}: A sequence model that considers other-dependency (i.e., only utterances of speakers $\mathcal{O} = \mathcal{U} \setminus u$ are used) and a temporal window of length $w$ in $\mathcal{C}_{\mathcal{O}}$ for \textbf{BiLSTM-PEC} or $\mathcal{C'}_{\mathcal{O}}$ for \textbf{DGCN-PEC}.


\medskip\noindent To summarize, any of these models is instantiated for each of the three types of sequence models ({\em emotion}, {\em text}, or {\em (emotion, text)}), and processed through either BiLSTM-PEC or DGCN-PEC.

\section{Experiments}
\label{sec:experiments}

\begin{table}[t!]
\centering
\small
\resizebox{\columnwidth}{!}{
\begin{tabular}{lccc} 
\toprule
Dataset & Type & \# Classes & \# Utterances\\
\midrule
\textsc{DailyDialog} & dyadic & 7 & 103.0k\\
\textsc{IEMOCAP} & dyadic & 11 & 6.8k\\
\textsc{MELD} & group & 7 & 13.7k \\
\textsc{Emotionlines:FRIENDS} & group & 8 & 14.5k \\
\bottomrule
\end{tabular}
}
\caption{Details of the datasets.}
\label{tab-data}
\vspace{-0.3cm}
\end{table}

\begin{table}[t!]
\centering
\resizebox{\columnwidth}{!}{%
\begin{tabular}{c>{\raggedleft\arraybackslash}p{1cm}>{\raggedleft\arraybackslash}p{1.5cm}>{\raggedleft\arraybackslash}p{1cm}>{\raggedleft\arraybackslash}p{1cm}>{\raggedleft\arraybackslash}p{1cm}>{\raggedleft\arraybackslash}p{1cm}>{\raggedleft\arraybackslash}p{1cm}} 
\toprule
&Orig. &$n \geq 1$  & $n \geq 2$ & $n \geq 3$ & $n \geq 4$ &$n \geq 6$ &$n \geq 8$\\  
\midrule
{\em neutral}  & 85572  &   73997 & 62907&52242&42127&26675&15639\\
{\em happiness} & 12885 &  11829 & 10464 &9212&7836&5590&3623\\
{\em surprise}  & 1823&  1708&  1436&1143&894&544&302\\
{\em sadness}   & 1150  &  1036 & 815&698&509&341&191\\
{\em anger}   & 1022  &  855 & 760&607&509&327&186\\
{\em disgust}   & 353  &  291  & 230&184&128&73&37\\
{\em fear}   &  174&  145 & 131 &104&87&57&34\\
\bottomrule
\end{tabular}
}
\caption{Emotion class label distributions on the reconstructed \textsc{DailyDialog}, for varying values of $n$.}
\label{table:2}
\vspace{-0.5cm}
\end{table}

\begin{table*}[!h]
\centering
\renewcommand{\arraystretch}{1.4}
\resizebox{0.99\textwidth}{!}{%

 \begin{tabular}{c|y yy |yyy|yy y|y yy|y y y|yy y|y yy|y y y } 

  \toprule
    Dataset &\multicolumn{6}{c|}{DAILYDIALOG}&\multicolumn{6}{c|}{IEMOCAP} &  \multicolumn{6}{c|}{MELD} & \multicolumn{6}{c}{FRIENDS} \\ 
 \toprule

    Model &\multicolumn{3}{c|}{BiLSTM-PEC}&\multicolumn{3}{c|}{DGCN-PEC} & \multicolumn{3}{c|}{BiLSTM-PEC} & \multicolumn{3}{c|}{DGCN-PEC} &\multicolumn{3}{c|}{BiLSTM-PEC} &\multicolumn{3}{c|}{DGCN-PEC} & \multicolumn{3}{c|}{BiLSTM-PEC} & \multicolumn{3}{c}{DGCN-PEC}\\ 

  \midrule
  \rowcolor{Gray1}
 Type &E & T& ET& E  & T& ET &E & T& ET& E  & T& ET&E & T& ET& E  & T& ET&E & T& ET& E  & T& ET \\  
 \midrule
 
1LB   &.20 &   .34 & .34 &\cellcolor{Gray2}  & \cellcolor{Gray2}  & \cellcolor{Gray2}    &.28 & .24  &   .25 &\cellcolor{Gray2} &\cellcolor{Gray2} & \cellcolor{Gray2}&.28 & .17&.19&\cellcolor{Gray1}&\cellcolor{Gray1}&\cellcolor{Gray1}& .27&.14&.18&\cellcolor{Gray1}&\cellcolor{Gray1}&\cellcolor{Gray1}\\
2LB  & \cellcolor{Green3}.45  &  .36 &\cellcolor{Green2}.41 & .44 &.37 &  .41 & \cellcolor{Green2} .38  &  \cellcolor{Green2}.33 & \cellcolor{Green2}.35 & .28& .23  &  .41& .30&\cellcolor{Green1}.21&.21&.30&.29 &.31 &\cellcolor{Red1}.27& \cellcolor{Green1}.19&.19&.28&.27&\cellcolor{Green1}.31\\
3LB  & \cellcolor{Red2}.44  &  .37& .42 & \cellcolor{Red2}.43  &.39  & .43&\cellcolor{Red1} .36 & .35 &.36& .31 & \cellcolor{Green2} .31 & .44 & \cellcolor{Red1}.28&.23&.22&\cellcolor{Green1}.34&.30&\cellcolor{Green1}.36& \cellcolor{Red1}.25&.21&.20&.28&.29&.34\\
4LB    &\cellcolor{Red1}.42 &  \cellcolor{Green1} .41 &\cellcolor{Red5}
 .42 & \cellcolor{Red4}.41    &\cellcolor{Green1}.44 & .46 & \cellcolor{Red2}.35 & .37 & \cellcolor{Red2}.35 & \cellcolor{Red3} .28 & \cellcolor{Green2}.41 & .45 & \cellcolor{Red2}.27&.24& .23& \cellcolor{Red4}.29&.31&.37 &\cellcolor{Red2}.24&.22 &.21& \cellcolor{Red1}.26&.30&.35\\
 5LB    &\cellcolor{Red1}.40 &   .42 & .43 & \cellcolor{Red1}.39    &.45 & \cellcolor{Red5}
.46& \cellcolor{Red2}.34 & .38 & \cellcolor{Red5}.35 & \cellcolor{Red2}.27 & .44 & \cellcolor{Red5}.45 & \cellcolor{Red2}.26 &.25&.24 & \cellcolor{Red2}.28&.34&.39 &\cellcolor{Red2}.23&.24&.22&\cellcolor{Red2}.25&.33&\cellcolor{Green1}.40\\
 \bottomrule
\end{tabular}%
}
\caption{Macro-average F1 scores of the  {\textbf {\em w}LB} sequence model, instantiated as any of the {\em emotion}, {\em text} and {\em (emotion, text)} sequence model type. Here DGCN-PEC outperforms BiLSTM-PEC as a classifier on the dyadic and group conversation datasets in the {\em text} and {\em (emotion, text)} sequence sequence model types. All { \em emotion} sequence model type trend negatively with more look backs shown in red highlighted table cells suggesting {\em recency}. }
\label{table:wLB}
\vspace{-0.5cm}
\end{table*}

In this section, we empirically evaluate the performance of our proposed models. We also examine the sensitivity of the models to the choice of the classifier, word embeddings, and the use of same/other speaker edges in our graph based model {DGCN-PEC}. Before presenting the results, we provide details of the datasets employed, the evaluation metric and the evaluation scenarios.


\smallskip\noindent\textbf{Datasets}. We use a number of existing conversation datasets in our experiments. Broadly, these datasets can be categorized as either being {\em dyadic conversations} (i.e., dialogues involving two speakers) or {\em group conversations} (i.e., dialogues involving multiple speakers). As dyadic conversations datasets, we use (i) \textsc{DailyDialog} \cite{li-etal-2017-dailydialog}, which consists of two-speaker dialogues pertaining to conversations about daily life, and (ii) \textsc{Iemocap} \cite{IEMOCAP}, which consists of dyadic sessions with actors performing emotional improvisations or scripted scenarios. For group conversations, we use (iii) \textsc{Meld} \cite{poria2018meld}, which  consists of multi-speaker dialogues from the comedy show Friends, and (iv) \textsc{Emotionlines:Friends} \cite{chen2018emotionlines}. The details of the datasets are summarized in Table~\ref{tab-data}.
 

\smallskip \noindent \textbf{Reconstructed datasets}. To train our sequence models that employ a temporal look back window of size $w$, we require that dialogues include at least a certain number of $n$ turns, where $n \geq w$. To accommodate for that we pre-process the original datasets and construct new ones that are subsets of the original datasets. For instance, if we are interested in predicting the emotion label $e_4$ at the fourth turn given the previous 3 turns (as in the example depicted in Table~\ref{tab:example}), then we have to extract all dialogues of the original dataset that are of at least four turns ($n \geq 4)$. This results in different distributions of the emotion class labels for the prediction problem. Table \ref{table:2} shows this effect for the case of the \textsc{DailyDialog} dataset, where the emotion class label distribution is given for varying values of $n$. Note that the order of turns is always preserved. If a conversation has fewer than $n$ turns, the entire conversation is discarded.

\smallskip\noindent\textbf{Evaluation Metric}. Due to the highly imbalanced nature of the datasets, the results are reported in terms of \textbf{macro-averaged F1 score}  that combines the per-class F1-scores into a single number, the classifier's overall F1-score. Recall that the F1-score for a single class is defined as $\frac{2 \times p \times r}{p + r}$, where $p$ and $r$ are precision and recall, respectively. For evaluation, the datasets are split into train and test sets with a 80/20 ratio. 

\smallskip\noindent\textbf{Evaluation Scenarios}. We seek answers to the following research questions: 
\begin{itemize}
\item Which of the three types of sequence models introduced ({\em emotion sequence}, {\em text sequence}, or {\em (emotion, text) sequence}) is more accurate?  In addition we evaluate the performance of a graph-based model such as {DGCN-PEC} and a sequential model such as {BiLSTM-PEC}.
\item What is the effect of incorporating self-dependency (vs. other-dependency) in the accuracy of the prediction model?
\end{itemize}

\subsection{Sequence Models Analysis}

\begin{table}[!t]
\centering
\resizebox{\columnwidth}{!}{%
\small
 \begin{tabular}{P{0.15\columnwidth}|P{0.11\columnwidth}P{0.11\columnwidth}P{0.11\columnwidth}|P{0.11\columnwidth}P{0.11\columnwidth}P{0.11\columnwidth}}
  \toprule
  Dataset &\multicolumn{3}{c|}{\textsc{DailyDialog}}
  &\multicolumn{3}{c}{MELD} \\ 
  \midrule
&E&T&ET&E&T&ET\\
 \midrule
2SLB&\textbf{.47}&.42&\textbf{.43}&.28&.30&.38\\
3SLB&.45&\textbf{.43}&.38&.28&\textbf{.36}&.41\\
4SLB&{.43}&.42&.36&\textbf{.29}&.33&\textbf{.43}\\
\midrule
2OLB&.43&.37&.40&.23&.24&.31\\
3OLB&.40&.30&.37&.25&.26&.34\\
4OLB&.35&.29&.35&.22&.30&.36\\
  \bottomrule
\end{tabular}%
}
\caption{Comparing the {\textbf {\em w}SLB} and {\textbf {\em w}OLB} sequence models in dyadic conversations, using \textsc{DailyDialog} and group conversations using \textbf{DGCN-PEC} on \textsc{meld} .}
\label{table:wSLB-wOLB}
\vspace{-0.5cm}
\end{table}

We evaluate the performance of the  \textbf{{\em w}LB} sequence models without incorporating any user dependency, instantiated for the three different types of sequences ({\em emotion sequence (E)}, {\em text sequence (T)}, {\em (emotion, text (ET))} sequence) and for varying values of the temporal look back (LB) window length $w=\{1, 2, 3, 4,5\}$ for \textbf{BiLSTM-PEC} and window length $w=\{ 2, 3, 4,5\}$  for \textbf{DGCN-PEC}, since  \textbf{DGCN-PEC} requires at least two utterances in the conversation sequence to create its graph. The results of our analysis are reported in Table~\ref{table:wLB} for dyadic and group conversations datasets. 

\noindent Looking at the overall trend across both dyadic and group conversations datasets, we observe that DGCN-PEC {\em text (T)} only and {\em (emotion, text (ET))} \textbf{{\em w}LB} sequence models outperform the {BiLSTM-PEC} models, which suggests that graph-based models which inherently incorporate user information yield better results than sequential models that do not incorporate any user information. 

In looking only at the dyadic conversations, we notice that for each dataset (\textsc{DailyDialog} and \textsc{Iemocap}), the best performance is obtained by {ET-DGCN-PEC} with 5LB. However, in taking a closer look, it seems that the best result of  \textbf{0.46} obtained by {ET-DGCN-PEC} with five look backs is comparable to that of \textbf{0.45} obtained by {E-BiLSTM-PEC} with just 2 look backs ($w=2$). From an efficiency point of view, for many real-time applications it is desirable to predict the next emotion using the least number of looks backs. Therefore, in these applications  using {E-BiLSTM-PEC} with 2 look backs ($w=2$) would be more efficient than using {ET-DGCN-PEC} with five look backs ($w=5$). 

In fact, the overall trends are highlighted in green and red, with green indicating an increasing pattern and red showing decreasing scores. A deeper shade of green depicts a higher value, and a deeper shade of red indicates a lower value. We notice that in the case of {E-BiLSTM-PEC} across both the dyadic datasets, the prediction based on more than 2 look backs ($w =\{3,4,5\}$) performs worse, suggesting that {\em recency} plays an important role in predicting conversation emotions. Since these experiments are performed on dyadic conversations, speakers $\mathcal{A}$ and $\mathcal{B}$ take turns of utterances. So, taking part in a conversation would look as follows: $\mathcal{A} \rightarrow \mathcal{B} \rightarrow \mathcal{A} \rightarrow \ldots$. Therefore, the prediction performance at two look backs($w=2)$ indicates that the emotion expressed in the utterance of the same speaker in which we attempt to predict for is likely an important factor for predicting the emotion expressed in the same speaker's next utterance. This also indicates the importance of {\em self-dependency}.


 We observe a similar trend across the two group conversation datasets where mostly, the smaller the number of look backs, the better the performance is, suggesting that {\em recency} plays an important role in group conversations as well.

\subsection{Incorporating Self-dependency} 

\begin{figure}[t!]
\centering
	\includegraphics[width=7.5 cm, height=5cm]{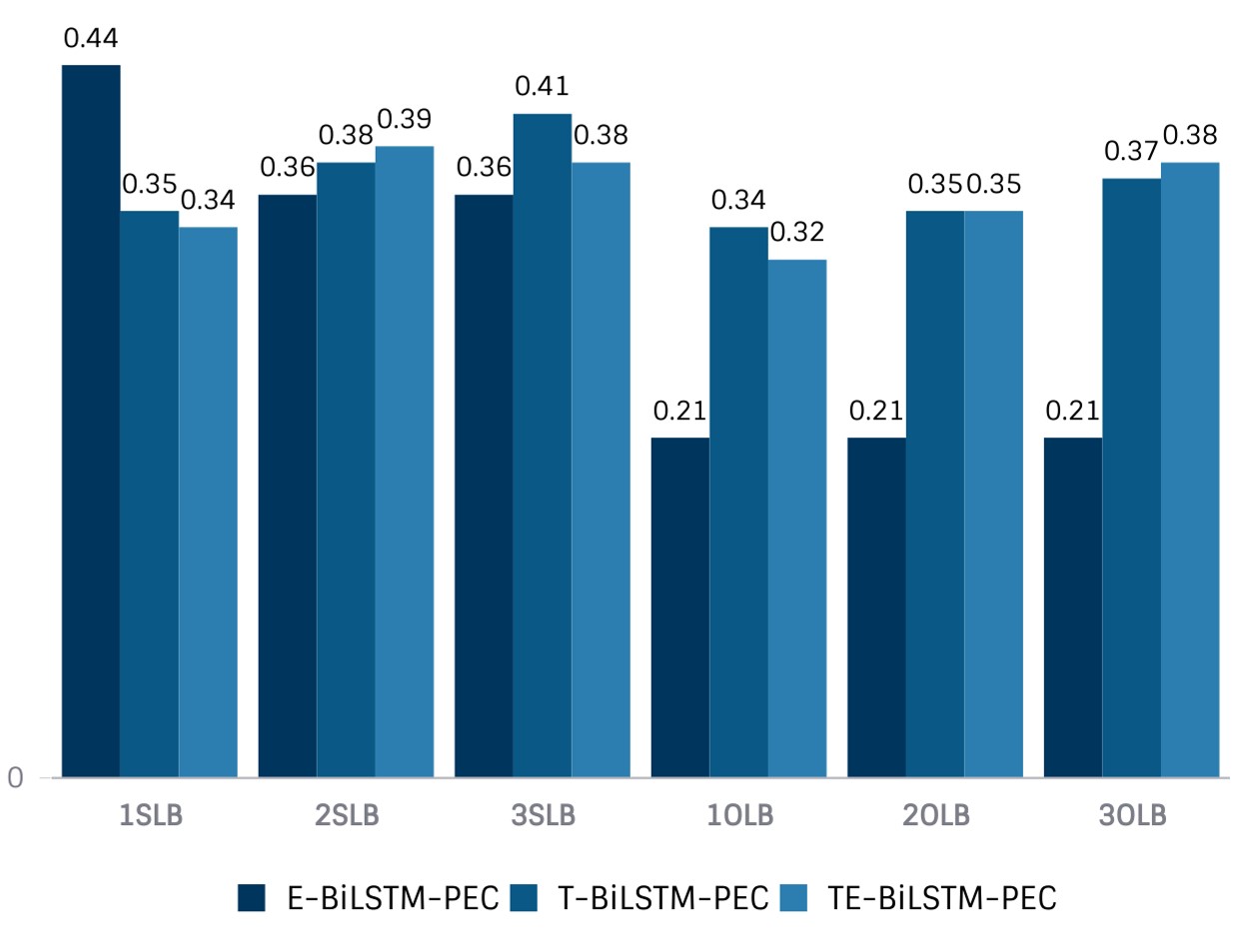}
	\caption{Comparing the {\textbf {\em w}SLB} and {\textbf {\em w}OLB} sequence models in \textsc{DAILYDIALOG} using \textbf{BiLSTM-PEC}.}
	\label{fig:chart3}
\vspace{-0.5cm}
\end{figure}

We evaluate the performance of the \textbf{{\em w}SLB} and \textbf{{\em w}OLB} sequence models, which incorporate {\em self-dependency} and {\em dependency on others}, respectively. We vary the values of the temporal look back window length $w=\{1, 2, 3\}$ for {BiLSTM-PEC} and $w=\{2, 3, 4\}$ for {DGCN-PEC}. 
First, we analyze the results presented in Table~\ref{table:wSLB-wOLB} for both dyadic and group datasets using \textbf{DGCN-PEC}. We notice that all the self-dependency models (\textbf{{\em w}SLB}) outperform all the other-dependency models (\textbf{{\em w}OLB}), for the same temporal window $w$. 
Next, we analyze the results presented in Figure~\ref{fig:chart3} for dyadic dataset using {BiLSTM-PEC}. Similar to DGCN-PEC, we notice that once again (\textbf{i}) all self-dependency models (\textbf{{\em w}SLB}) consistently outperform all other-dependency models (\textbf{{\em w}OLB}), for the same temporal window $w$, and (\textbf{ii}) the best overall performance is achieved by \textbf{1SLB} using the {\em emotion sequence}, thus providing further support to the importance of {\em recency} and {\em self-dependency} when predicting emotions in conversations.

\begin{figure}[t!]
   \includegraphics[width=7.5 cm, height=5cm]{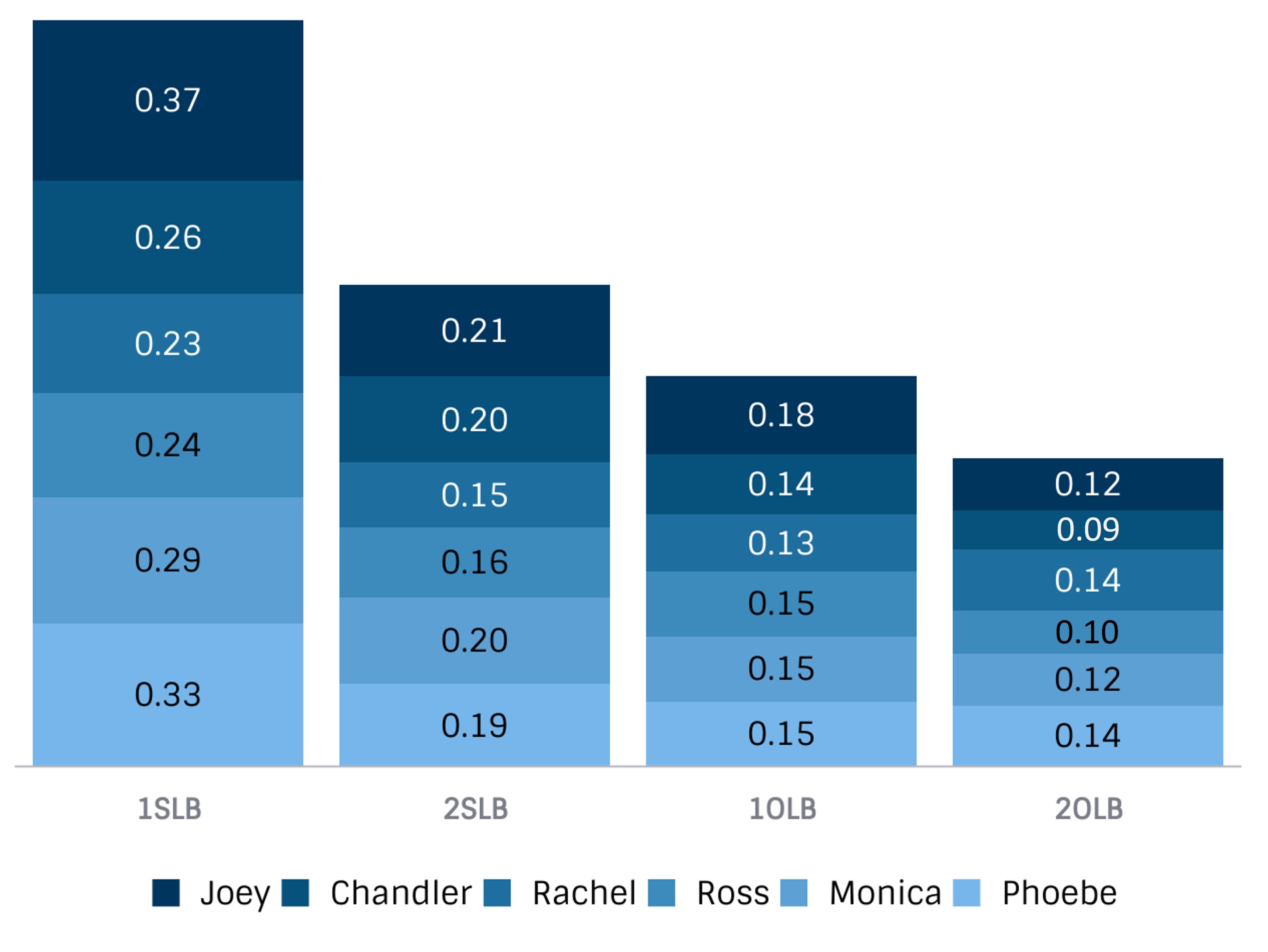}
   \caption{Macro-avg F1 score using \textbf{E-BiLSTM-PEC} for the MELD group conversation dataset. Each color represents one of the six characters in the dataset.}
   \vspace{-0.5cm}
   \label{fig:MELD-users}
\end{figure}

In Figure~\ref{fig:MELD-users}, we study the performance of {BiLSTM-PEC} on a group dataset {MELD} consisting of dialogues and utterances  from the popular TV series Friends featuring six characters: {\em Rachel}, {\em Monica}, {\em Phoebe}, {\em Ross}, {\em Chandler} and {\em Joey}. We observe that: 
(\textbf{i}) for any character, all self-dependency models (\textbf{{\em w}SLB}) outperform the other-dependency models (\textbf{{\em w}OLB}), for the same temporal window $w$, thus providing further support to the importance of the {\em self-dependency} aspect when predicting emotions in conversations.(\textbf{ii}) for any character, the best overall performance is achieved by the \textbf{1SLB} model compared to \textbf{2SLB}, thus providing further support to the importance of {\em recency} aspect when predicting emotions in conversations. 


\begin{figure*} 
	\centering
		\begin{subfigure}{0.28 \textwidth}
			\includegraphics[width=\linewidth]{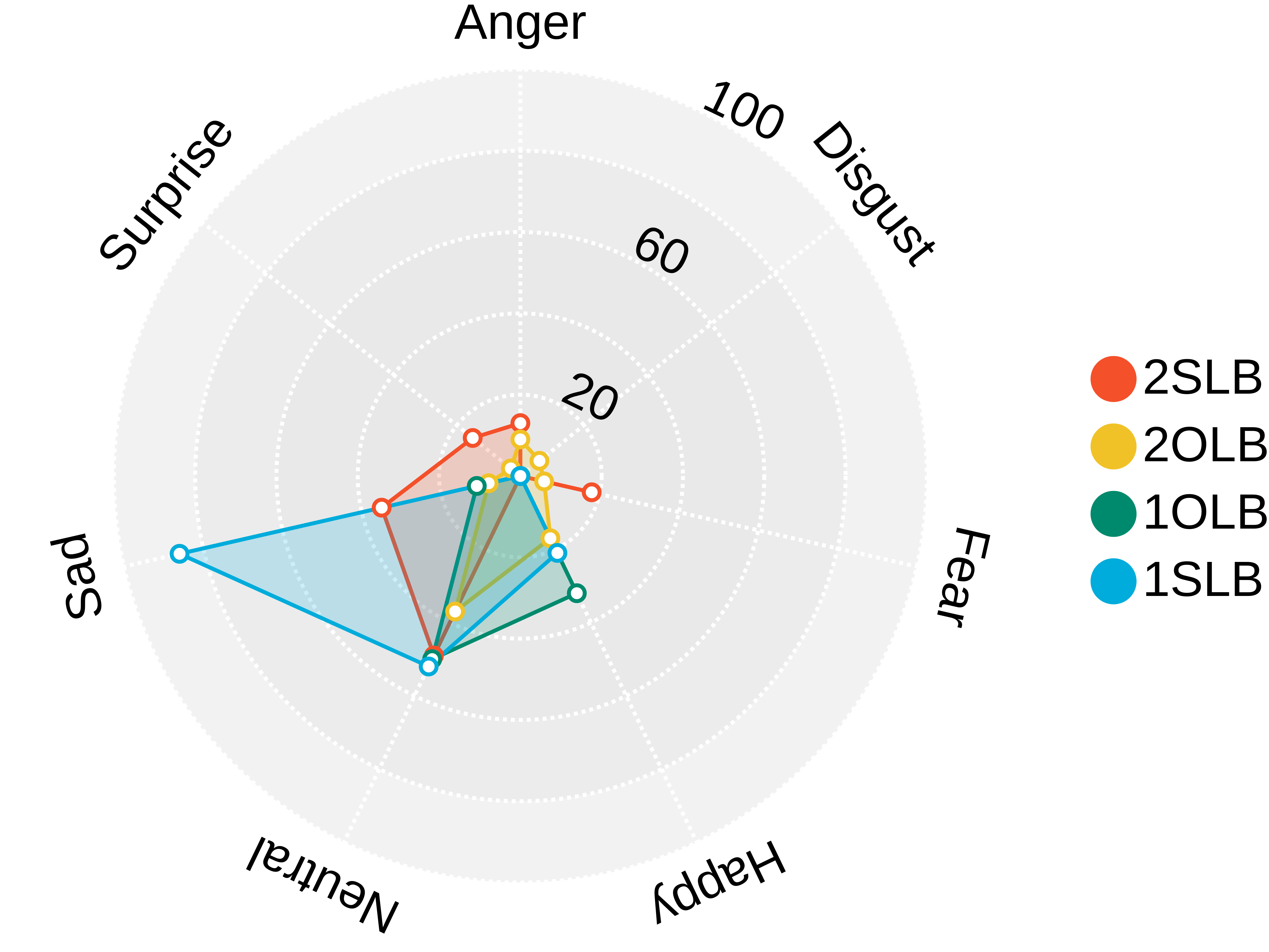}
			\caption{Chandler}
			\label{Chandlerprofile}
		\end{subfigure}
		\begin{subfigure}{0.28 \textwidth}
			\includegraphics[width=\linewidth]{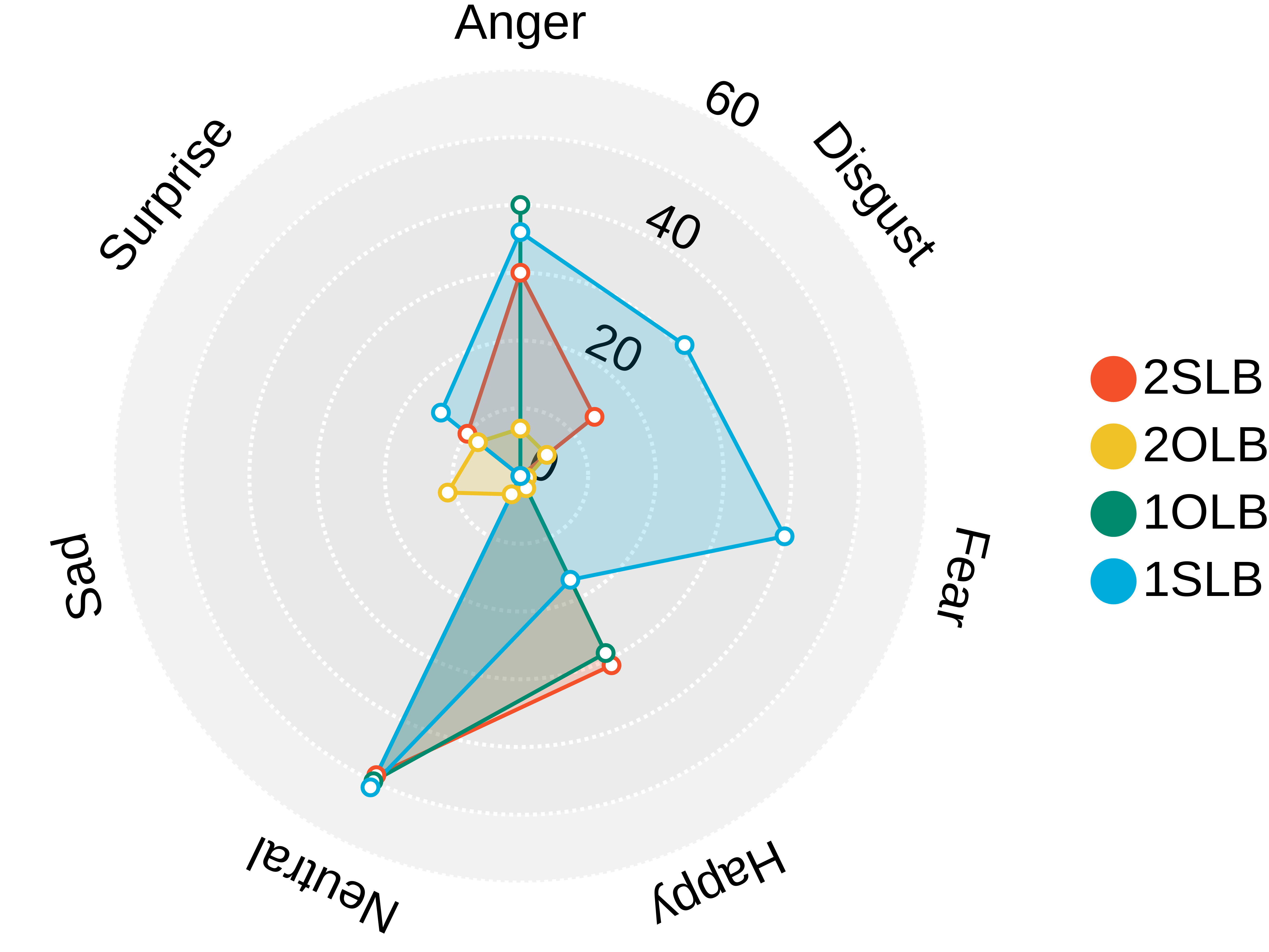}
			\caption{Joey} 
			\label{Joeyprofile}
		\end{subfigure}	
		\begin{subfigure}{0.28 \textwidth}
			\includegraphics[width=\linewidth]{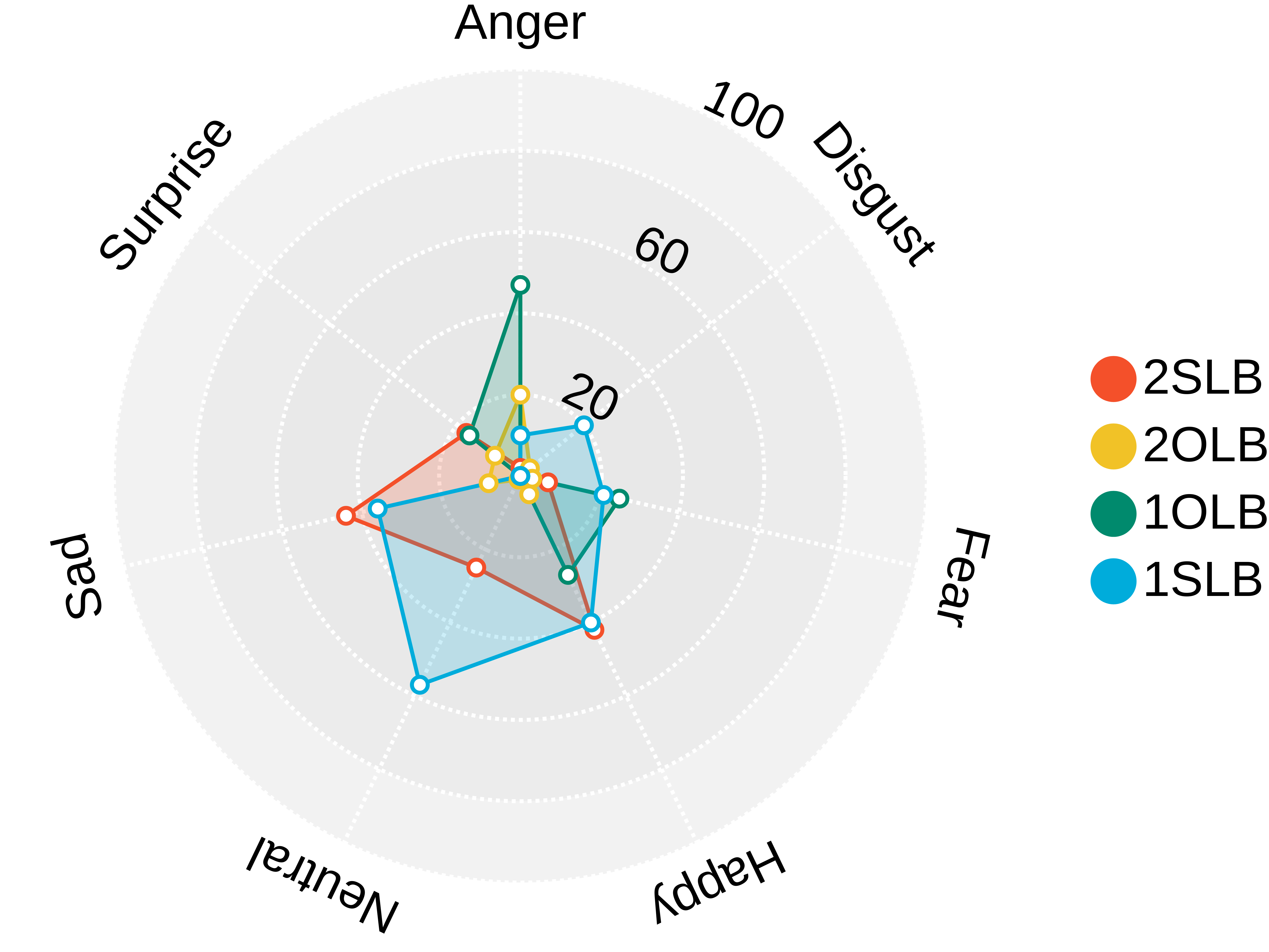}
			\caption{Phoebe} 
			\label{Pheobeprofile}
		\end{subfigure}			
		\caption{Emotion profiles of three example characters ({\em Chandler, Joey, Pheobe}) coming from the \textsc{MELD} dataset.}
		\label{fig:MELD-profiles}
		\vspace{-0.5cm}
	\end{figure*}



We further refine the analysis for each of the six characters in the \textsc{MELD} group conversation dataset to explore the effect of the model by each of the emotion categories ({\em neutral}, {\em sad}, {\em surprise}, {\em anger}, {\em disgust}, {\em fear}, {\em happy}). Figure~\ref{fig:MELD-profiles} shows the radar plots of three of the main characters ({\em Chandler, Joey and Pheobe}) and their emotion signals. The plots of the other three characters ({\em Rachel, Ross and Monica}) are similar (see Appendix). Again, we observe that for any character, the (\textbf{{\em w}SLB}) models (\textbf{1SLB}, \textbf{2SLB}) outperform the (\textbf{{\em w}OLB}) models (\textbf{1OLB}, \textbf{2OLB}) --- see the larger area occupied by the self-dependent and more recent models. At the same time, the models also seem to be capturing the typical semantics of each character's profile.

\subsection{Ablation Study}

\subsubsection{Varying past window size, $pw$}
Recall that in Table~\ref{table:wLB}, we presented the results where the overall best performance for each dataset was obtained using {ET-DGCN-PEC} with five look backs ($w=5$), and that the past window $pw$ variable used for creating the edges in {DGCN-PEC} was empirically set to $pw=3$ capturing only recent data around any utterance. In Table~\ref{past-window} we present a detailed analysis varying the values of $pw$  from  $pw=\{3,4,5\}$,  and note that increasing $pw$, i.e., increasing  history data, gives worse results, confirming the importance of  {\em recency}. 


\begin{table}[!t]
\centering
\resizebox{\columnwidth }{!}{
\begin{tabular}{p{0.4\columnwidth}P{0.3\columnwidth}P{0.3\columnwidth}P{0.3\columnwidth}}
\toprule

Dataset & $pw=3$& $pw=4$& $pw=5$\\
\midrule
\textsc{DailyDialog} & \textbf{.46} & .39& .39 \\
\textsc{IEMOCAP} & \textbf{.45} &.38 & .40 \\
\textsc{MELD} & \textbf{.39} & .34 &.36 \\
\textsc{Friends} & \textbf{.40}& .34 &.35\\

\bottomrule
\end{tabular}
}
\caption{Sensitivity analysis of the past window size parameter $pw$.}
\label{past-window}
\vspace{-0.5cm}
\end{table}

\subsubsection{Varying speaker edges}




We further analyze the role of {\em self-dependency} by creating another variant of {DGCN-PEC} denoted as \textbf{DGCN-PEC-S} that uses only the same speaker edges in the graph (i.e., only use edges of type $ u_i \rightarrow u_i$ for $i$ in the range of speakers). Table ~\ref{table:DGCN_S} summarizes the results and shows that on average for the {\em(text (T))} only and the combined {\em(emotion, text (ET))} sequences, the results are either higher for {DGCN-PEC-S} or the same for both {DGCN-PEC} and {DGCN-PEC-S}, suggesting that it is both  more efficient and accurate to use speaker edges only. This further confirms the importance of {\em self-dependency} in predicting future evoked emotion. 

\begin{table}[!t]
\resizebox{ \columnwidth}{!}{%
\centering
 \begin{tabular}{c|ccc |ccc|ccc|ccc} 

  \toprule
 &\multicolumn{6}{c|}{\textsc{DailyDialog}} &\multicolumn{6}{c}{MELD} \\ 
\midrule

   &\multicolumn{3}{c|}{DGCN-PEC}&\multicolumn{3}{c|}{DGCN-PEC-S} &\multicolumn{3}{c|}{DGCN-PEC} &\multicolumn{3}{c}{DGCN-PEC-S} \\ 
     &E& T& ET& E& T& ET&E& T& ET&E& T& ET \\ 
  \midrule
 \midrule
2LB&.44& .37& .41 &\textbf{.46}&.38&.41&.30&.29&.31&.28&.30&\textbf{.33}\\

3LB&\textbf{.43}& .39& .40&\textbf{.43}&.39&.42&.34	&.30&.36&.31&.32&\textbf{.36}\\
 
4LB&.41& .44& \textbf{.46}&.40&.44&\textbf{.46}&.29	&.31&\textbf{.37}&.26&.32&\textbf{.37}\\

5LB &.39& .45& .46&.37&.46&\textbf{.47}&.28&.34&\textbf{.39}&.28&.33&.38\\

 \bottomrule

\end{tabular}%
}
\caption{Comparing \textbf{DGCN-PEC} to \textbf{DGCN-PEC-S} in dyadic conversations, using \textsc{DailyDialog} and in group conversations, using \textsc{MELD}.}
\label{table:DGCN_S}
\vspace{-0.5cm}
\end{table}

\section{Conclusions}
\label{sec:conclusions}

We introduced the novel problem of predicting evoked emotions in conversations ({\em PEC}) and proposed two novel neural network models to address it -- BiLSTM-PEC and DGCN-PEC. We proposed three modeling dimensions relevant to this task and conducted an extensive empirical analysis to determine the effect of {\em recency} and {\em self-dependency} on a model's prediction accuracy. Our results indicate that 
({\em i}) for ({\em text}) and ({\em emotion,text}) utterances, DGCN-PEC, the {\em graph network model}, that inherently accounts for user information and recency outperforms BiLSTM-PEC, the {\em sequence model}; 
({\em ii}) using same speaker data  (and/or same speaker edges) further improves the results of DGCN-PEC, confirming the role of {\em self-dependency} in emotion prediction; and
({\em iii}) for { \em emotion} sequences, the BiLSTM-PEC model that uses only same user data and as few as 2 lookbacks, performs similarly to the more complicated DGCN-PEC model needing at least 4 lookbacks. A simpler model that can predict emotions with less lookbacks may be more efficient for certain applications.
In conclusion, when designing emotion prediction models, taking into consideration the dimensions of {\em recency} and {\em self-dependency} seems to be beneficial.

\section*{Ethical Considerations}

Research that attempts to infer or predict user emotions should certainly be used in a responsible and transparent manner with proper explicit consent of the user. Moreover, not relying on any protected class information (directly or indirectly) may further ensure that the models do not exploit any underlying biases of the system. In this work, we use publicly available datasets, so it is possible that the biases exhibited in the existing datasets are reflected in our supervised models. Although we develop this model with a positive intention in mind, that of facilitating positive outcomes such as pre-emptive toxicity detection in social media forums, unfortunately, there is potential for such models to be misused in unexpected purposes such as for obfuscating toxic or hate speech. 



\bibliography{acl2020}
\bibliographystyle{acl_natbib}
\appendix

\newpage

\appendix

\section {Computational considerations}
For each one of the experimental results in this work, the datasets are split to 80\% training and 20\% testing. We run each result exactly 30 epochs and report the maximum value.
The experiments are run on Dell  Alienware Aurora R7 desktop with a Nvidia 1080Ti RTX Graphic card.

\section {Implementation}
In this work, we utilize a number of pre-existing packages. For our  textual input, we specifically use Glove's 840B token and 300d vector embeddings. Pre-processing the text input is done using (i) Keras tokinzer and sequence padding, and (ii) NLTK stopwords and lemmatizer. For Metric reporting we use Sklearn metrics. For our models and neural network layers we use Keras and Torch. 

When running DGCN-PEC, the parameters for cuda and nodal attention are set to False. All other parameters, unless stated otherwise in the paper, use default values of the original implementation of DialougeGCN. 
\section{Handling imbalanced classes}
For each of the datasets, the class distribution  was examined to determine the nature of the dataset. Typically, the label distribution in emotion datasets is quite imbalanced, and that remains true for these conversation datasets labeled with emotion categories as well.


 As an initial study using BiLSTM-PEC, for the DAILYDIALOG dataset we experiment with three strategies for handling the imbalanced data. 
 
 \noindent (i) \textbf{Oversampling (OS)}: All the minority classes are over-sampled to match the support of the majority class (i.e., {\em neutral}) using sampling with replacement. 
 
 \noindent (ii) \textbf{Class weights (CW)}: Assuming $L={l_1,...,l_k}$ to be the set of all possible emotion classes, where $|l_i|$ is the number of samples in class $l_i$, we assign weights to each of the classes as $ CW(l_i) = \frac{|L|}{|l_i|}$. 
 
 \noindent (iii) \textbf{Smooth weights (SW)}: The class weights can be further smoothed by defining $score(l_i) = log(\mu\frac{| L|}{|l_i|})$ ($\mu=0.15$ for our experiments), and $SW(l_i)=\max(score(l_i), 1)$.




\begin{figure}[h]
   \includegraphics[width=\columnwidth]{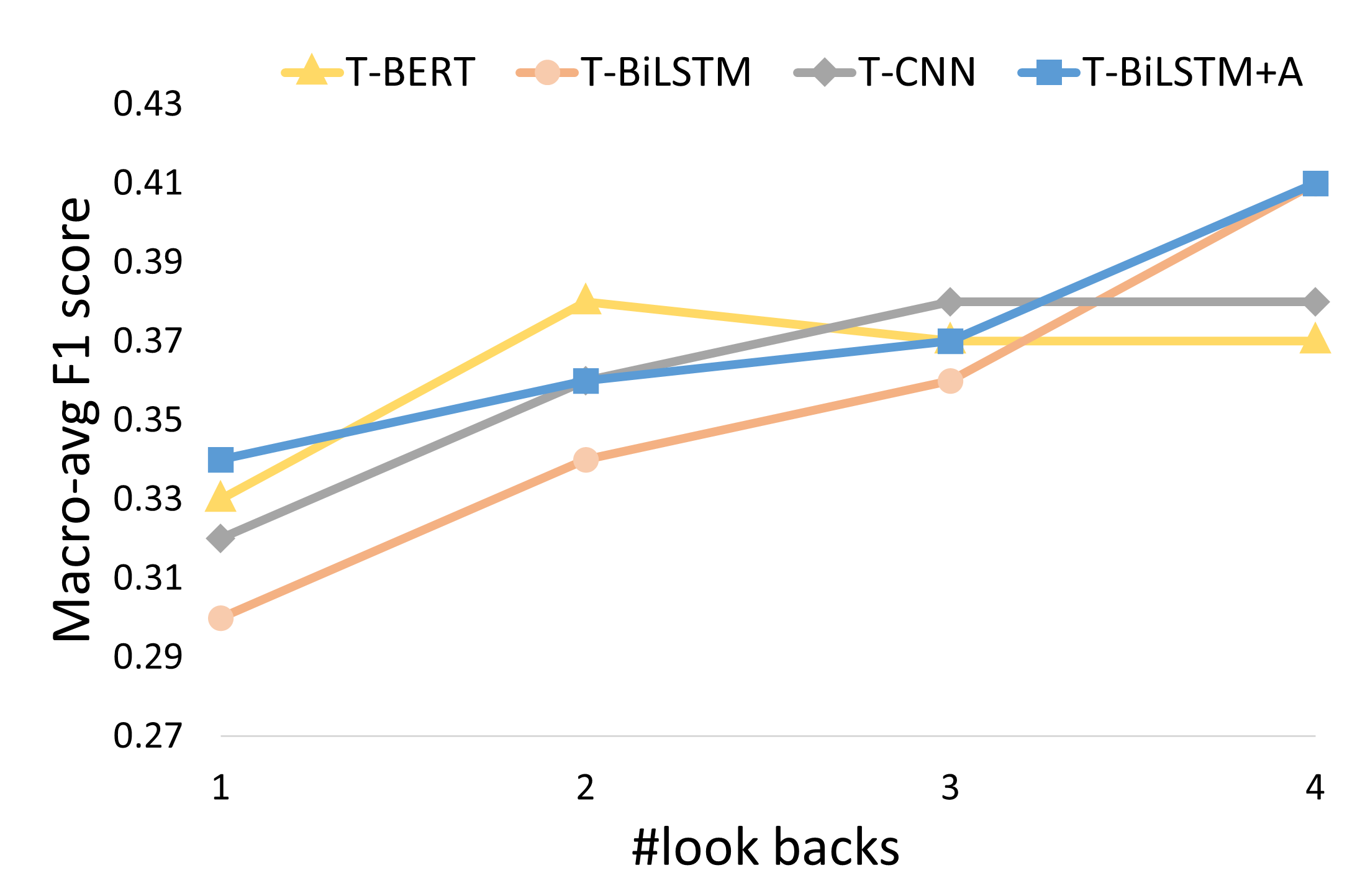}
   \caption{Comparing different sequence classifiers  with {\textbf {\em w}LB} and text sequences for \textsc{DailyDialog}.}
   \label{classifier}
\end{figure}
\begin{figure}[h]
   \includegraphics[width=\columnwidth]{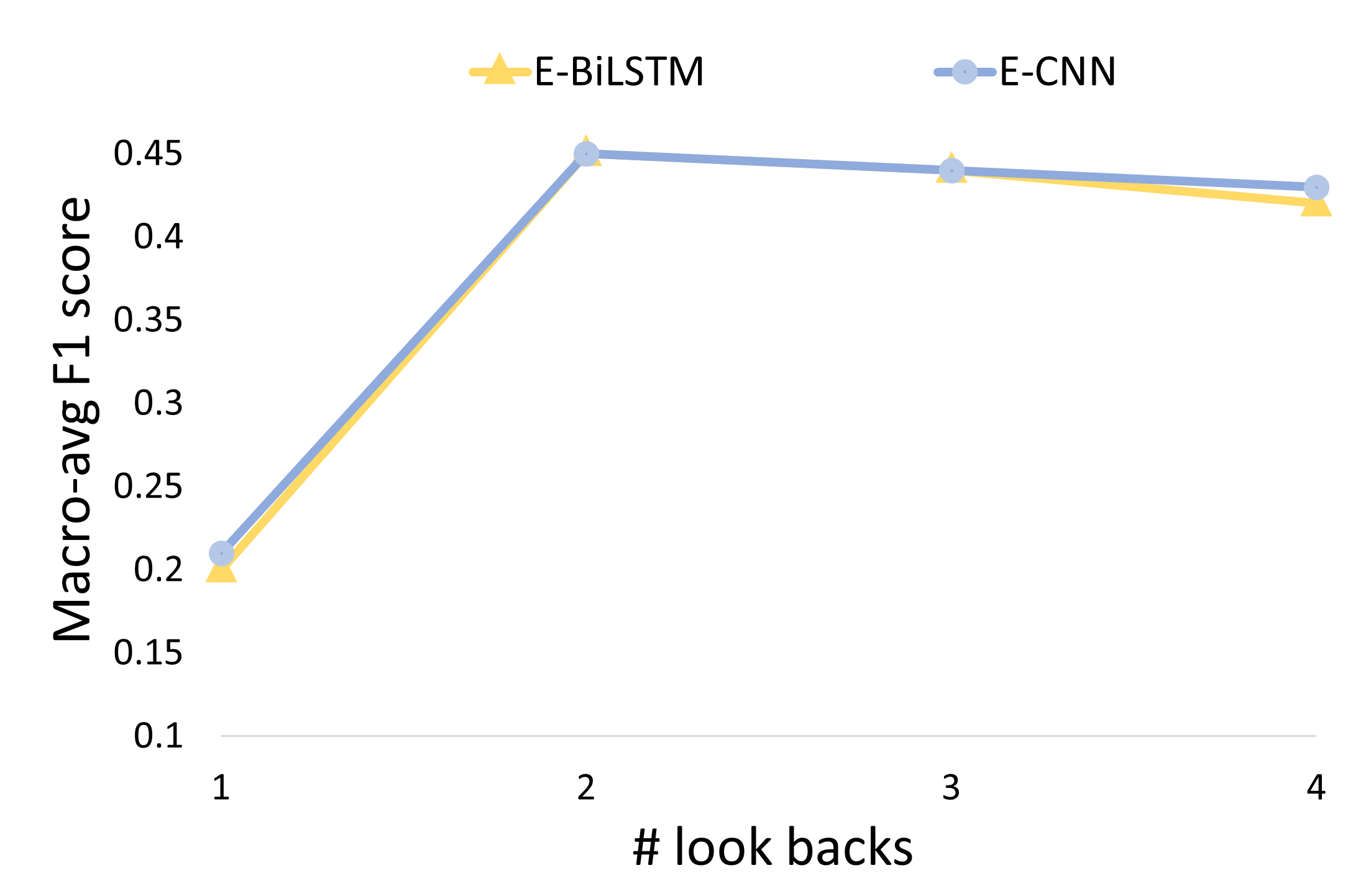}
   \caption{Comparing classifiers  {\textbf {\em w}LB} in the Emotion model using  in DAILYDIALOG. With the exception of T-BERT, all classifier use Glove embedding. }
   \label{classifier2}
\end{figure}

	\begin{figure*}[t] 
	\centering
		\begin{subfigure}{0.32\textwidth}
			\includegraphics[width=\linewidth]{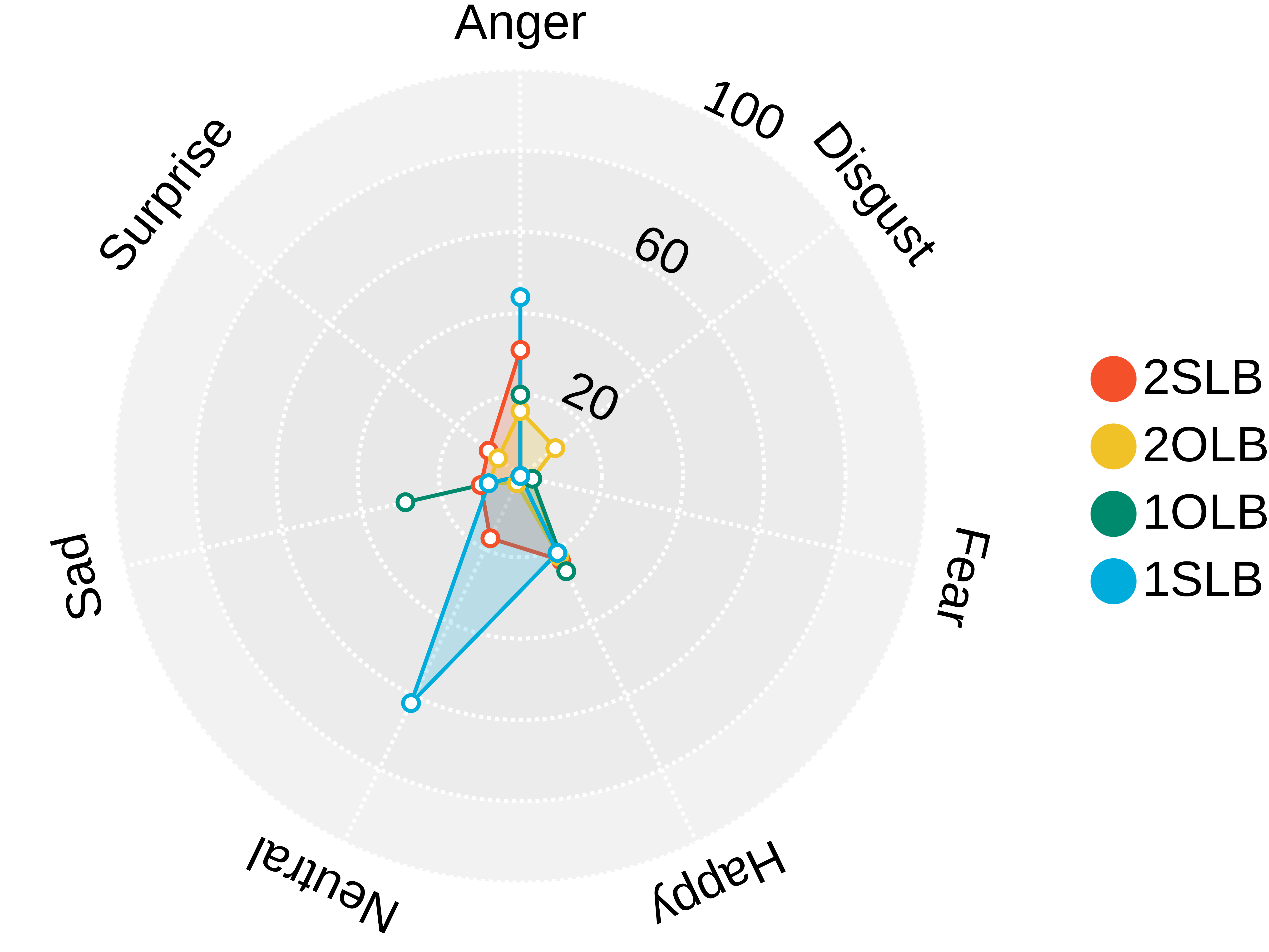}
			\caption{Ross} 
			\label{Rossprofile}
		\end{subfigure}
		\bigskip
		\begin{subfigure}{0.32\textwidth}
			\includegraphics[width=\linewidth]{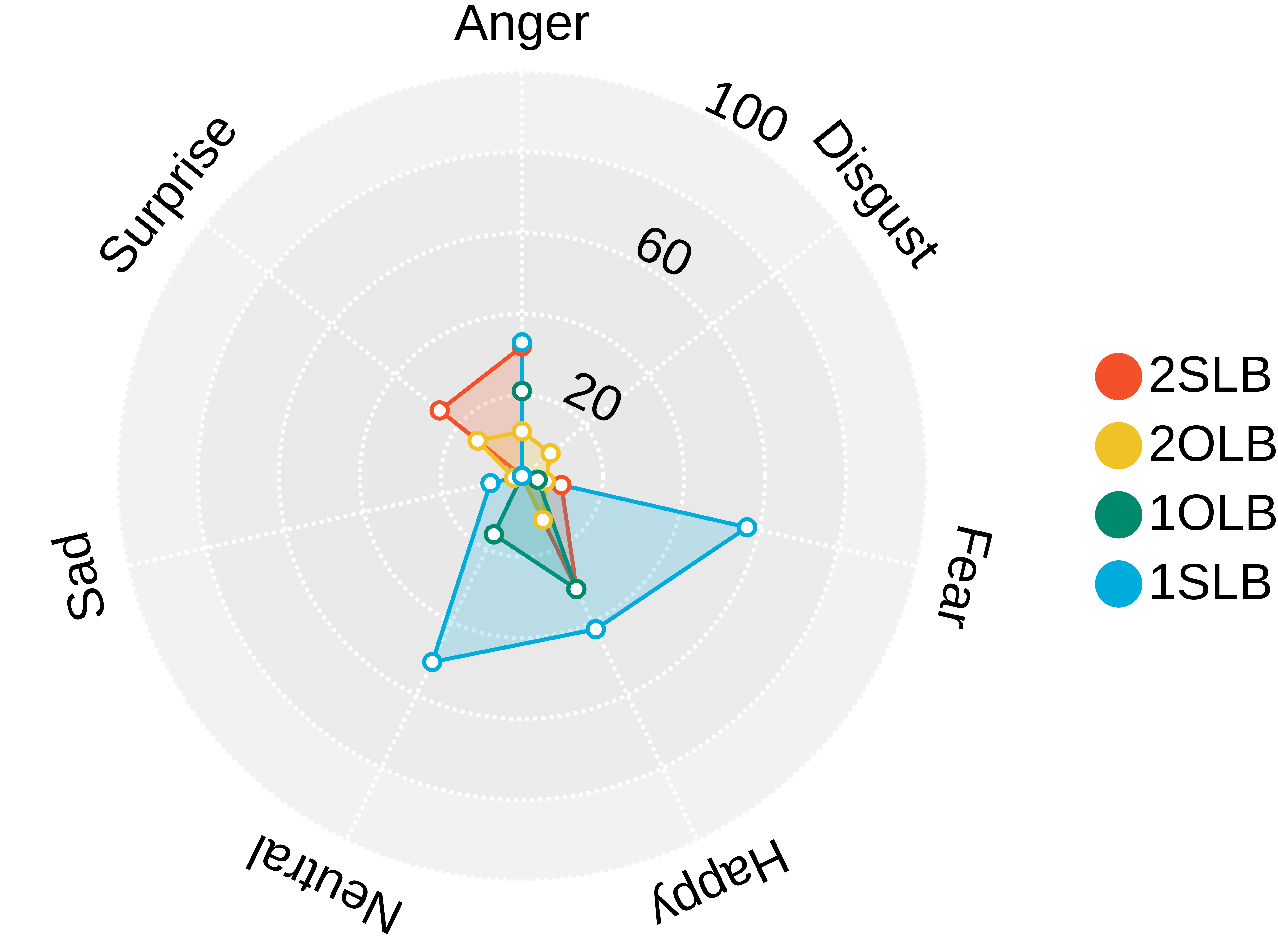}
			\caption{Monica} 
			\label{Monicaprofile}
		\end{subfigure}
		\begin{subfigure}{0.32 \textwidth}
			\includegraphics[width=\linewidth]{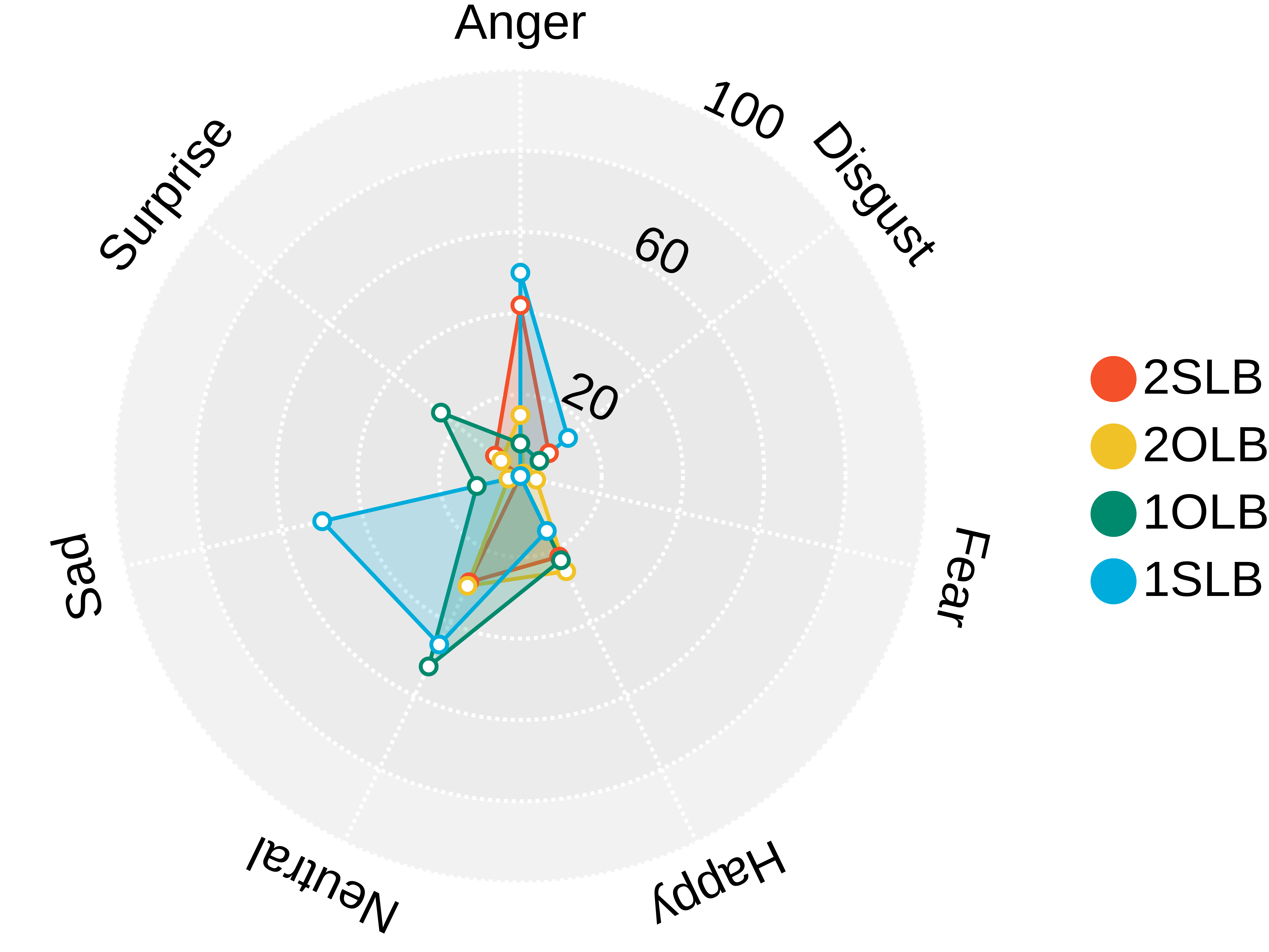}
			\caption{Rachel} 
			\label{Rachelprofile}
		\end{subfigure}
			
		\caption{User profiles derived from the emotion sequences of the \textsc{MELD} dataset for the characters.}
		\label{fig:profiles}
	\end{figure*}

\begin{figure}[h]
   \includegraphics[width=\columnwidth]{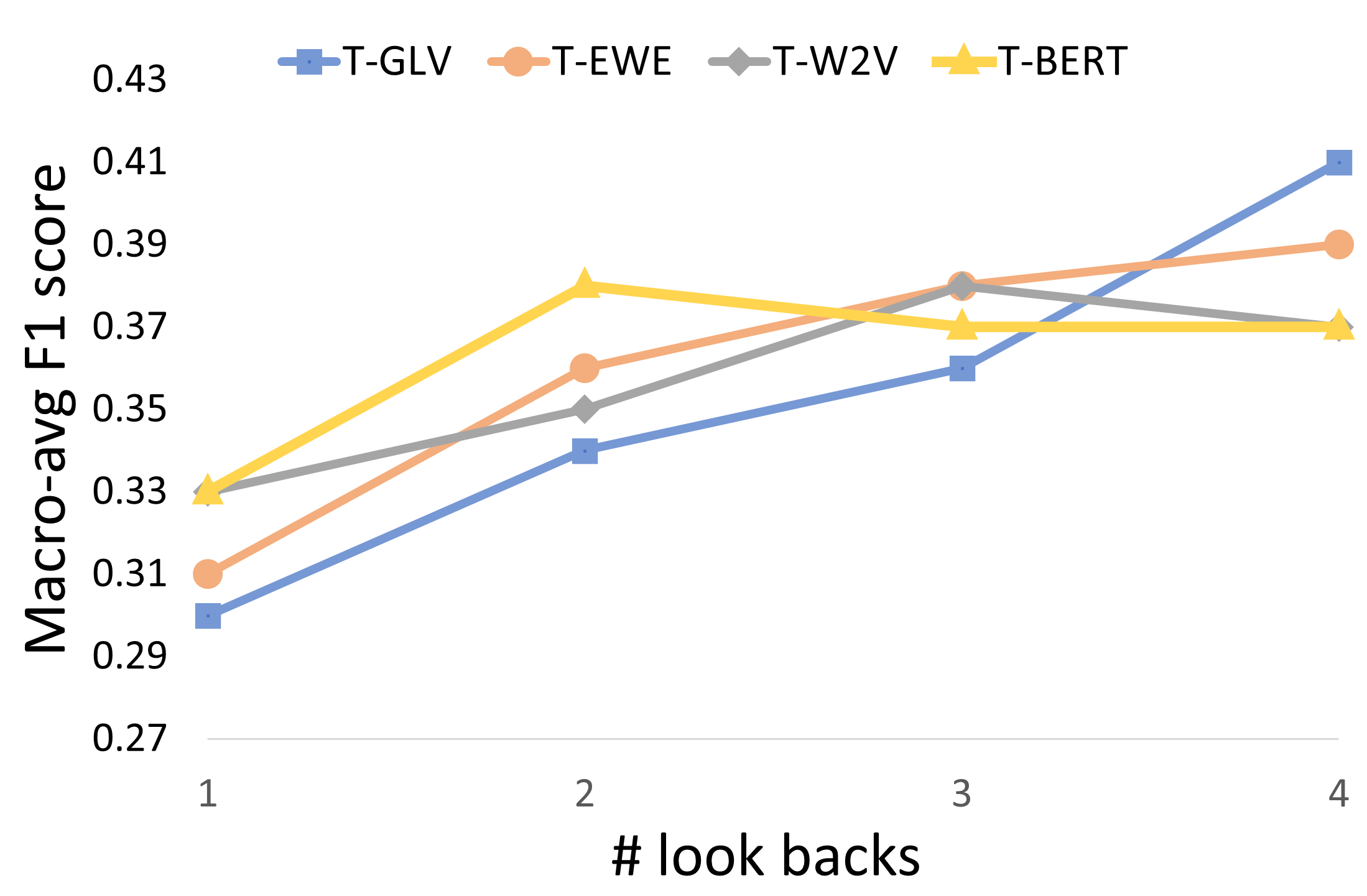}
   \caption{Comparing {\textbf {\em w}LB} using BiLSTM-PEC in the  text model  with pre-trained embeddings including GloVe, word2vec, EWE and BERT for \textsc{DailyDialog} dataset.}
   \label{fig1}
   \vspace{-0.5cm}
\end{figure}

\smallskip\noindent The results of applying the various strategies including no balancing (NB) are presented in \ref{table:handle}.  We observe that  balancing the dataset yields improvement over leaving it as is, especially as the length of the temporal window $w$ increases, with the best consistent performance obtained by applying smooth weights (SW), which is the method used in the remainder of the experiments.  Table \ref{table:handle}provides an extensive breakdown of the handling of imbalance in the data of one of the datasets, \textsc{DAILYDIALOG}. Identical experiments were conducted on each of the data sets to determine the best possible strategy to use.

\begin{table*}[!h]
\resizebox{\textwidth}{!}{%
\centering
 \begin{tabular}{||c|c| c|c |c |c|c|c |c |c| c|c |c |c| c|c |c ||} 
   \hline

  \hline
    wLB &\multicolumn{4}{c|}{1LB}&\multicolumn{4}{c|}{2LB} &  \multicolumn{4}{c|}{3LB} & \multicolumn{4}{c||}{4LB} \\ 
 \hline
  \hline
 Method &NoB & SW & CW & OVS  &NoB  & SW & CW & OVS &NoB  & SW & CW & OVS &NoB & SW & CW & OVS\\  
 \hline\hline
Neutral   &.92  &   .92 & .90 &.90  & .91 &   .92 & .89 &.89 & .91  &   .91 & .85 &.85 & .92  &   .91 & .84 &.85\\
 \hline
Anger  &0  &  0 &.14 & .14 &.56 &  .56 &.43 & .43 &.56  &  .54 &.41 & .28& 0  &  .49 &.37 & .36\\
 \hline
Disgust  &0  &  0& .03 &.03  &0  & .48 & .35 & .35 &.05& .43 & .31 & .35  & 0 & .41 & .29 & .29\\
 \hline
Fear    &0 &   0 & 0 & 0    &0 & .29 & .22 & .26 & 0 & .29 & .24 & .25 & 0  &   .34 & .28 & .21  \\
 \hline
Happiness   &.54   &  .54 & .54 & .54  &.49  &  .49 & .57 & .57 & .55   &  .55 & .56 & .56 & .54 &  .45 & .54 & .54\\
 \hline
Sadness   &0  &   0  & .07& .07  &0  &.28  & .22& .23& 0 &   .38  & .29 & .29 & 0 &   .32  & .2 & .16\\
 \hline

Surprise   &0   &   0 & 0& 0   &0  &   0 & .13& .14  & 0&  0 & .07 & .08 & 0  &   .01 & .06 & .07\\
 \hline
  \rowcolor{Gray2}
 macro avg   &.21 & .21 & .24 & .24 &.28  & .43 & .40 & .41& .30 & .44 & .39 & .38& .21  & .42 & .37 & .35 \\
 \hline
  \rowcolor{Gray1}
 loss  & .53& .55 & 1.83 & 1.82  &.523 & .5 & 1.54 & 1.52 & .491 &.5 & 1.53 & 1.56 & 0.548  & .49 & 1.5 & 1.44 \\
 \hline

\end{tabular}%
}

\caption{\textbf{Handling imbalanced data:} The Macro-average F1 score on  {\em emotion(E)} sequences in the \textsc{DAILYDIALOG} dataset with wLB where $w=\{1,2,3,4\}$ , where NoB= No-Balancing, SW = Smoothen-Weights, CW = Count-Weights and OVS = OVer-Sampling.  }
\label{table:handle}
\end{table*}

\section {Speaker Emotion Profiles}
 To examine speaker dependency in group conversation datasets, data related to each speaker is extracted. For each speaker we conduct the same set of experiments. we analyze {\em recency} and {\em self-dependency} on the characters in \textsc{MELD} dataset. Figure~\ref{fig:profiles} shows the results of the analysis for three of the main character  ({ \em Rachel, Ross and Monica}) and each emotion signal using radar plots. The other three characters ({\em Chandler, Joey and Pheobe}))are listed in the paper.we observe that: (\textbf{i}) for any character, the (\textbf{{\em w}SLB}) models (\textbf{1SLB}, \textbf{2SLB}) outperform the (\textbf{{\em w}OLB}) models (\textbf{1OLB}, \textbf{2OLB}).See the larger area occupied by the self-dependent and more recent models. At the same time, they manage to properly capture the semantics of each character's profile.

\section {Comparing Classifiers}

When comparing between  different classifiers using the {\em text} sequence, we substitute BiLSTM with attention denoted as BiLSTM+A  in our sequential neural net model with one of the following classifiers: BiLSTM, CNN and BERT. In general, the results of all the classifiers  fall within a narrow range albeit with varying trends as seen in Figure~\ref{classifier}. Exploring such trends further may be possible when more look backs are available. Unsurprisingly, BiLSTM with attention is consistently better than BiLSTM. Another interesting observation is the emotion series similar behaviour given any classifier, shown in Figure\ref{classifier2}.

\section {Comparing  Word Embeddings}
For initializing the embedding layer of the {\em text} (T) sequence classifiers, we experimented with four types of pre-trained word embeddings including word2vec \cite{mikolov2013distributed}, GloVe \cite{pennington2014glove}, EWE \cite{agrawal2018learning}, and BERT (base and uncased) \cite{devlin2018bert}. The results of choosing different embeddings as tested on \textsc{DailyDialog} are shown in Figure~\ref{fig1}. Notably, we observe that there is no consistently best word embeddings, and therefore, we choose GloVe representations for all the experiments.


\end{document}